# ARYA: A Physics-Constrained Composable & Deterministic World Model Architecture

Seth Dobrin, PhD
CEO | ARYA Labs
seth@aryalabs.io

Lukasz Chmiel
CTO | ARYA Labs
lukasz@aryalabs.io

## Abstract

This paper presents ARYA, a composable, physics-constrained, deterministic world model architecture built on five foundational design principles: nano models, composability, causal reasoning, determinism, and architectural AI safety. We demonstrate that the platform satisfies the formal requirements established by the world model research community, including state representation, dynamic prediction, causal and physical awareness, temporal consistency, generalization, learnability, and applicability to planning and control. Unlike monolithic large language models, ARYA implements these capabilities through a hierarchical system-of-system-of-systems of specialized nano models orchestrated by AARA (ARYA's Autonomous Research Agent), an always-on cognitive daemon that operates a continuous sense-decide-act-learn loop. The nano model architecture provides linear scaling, sparse activation (invoking only task-relevant models), selective untraining, and sub-20-second training cycles. Combined, these properties resolve the traditional tension between capability and computational efficiency. A central contribution is the "Unfireable Safety Kernel", an architecturally immutable safety boundary that cannot be turned off, bypassed, or circumvented by any component of the system, including its own self-improvement engine. This layer is not a statement on social or ethical alignment; rather, it is a technical framework for ensuring that human control and governance are maintained as the system's autonomy increases. Safety is not a policy layer applied after the fact; it is an architectural constraint that governs every operation the system performs. We present the formal alignment between ARYA's architecture and the canonical world model requirements, summarize its state-of-the-art performance across 6 of 9 competitive benchmarks, and describe its deployment across seven active industry domain nodes (aerospace, pharma manufacturing, oil & gas, smart cities, biotech, defense, and medical devices), and report empirical evaluation on nine external benchmarks where AARA achieves state-of-the-art on six—including causal reasoning, physics reasoning, PhD-level science, enterprise workflows, embodied planning, and AI safety—with zero neural network parameters).

## 1. Introduction

The development of AI systems that can reason about the world, predict the consequences of actions, and continuously improve their own capabilities represents one of the central challenges in computer science. At the core of this challenge is the pursuit of world models: internal representations that enable an agent to simulate the environment's dynamics, predict future states, and plan actions without direct interaction with the real world [1] [2] [3]. A system that genuinely understands the dynamics of its operating environment, and one that merely correlates patterns in training data while modeling causal structure, physical constraints, and temporal evolution, has the foundation for autonomous reasoning, planning, and self-improvement.

The ARYA platform is ARYA Labs' answer to this challenge. It is a governed, safety-first framework built around a composable world model architecture that serves multiple industry domains. The system also features self-improvement, cross-domain generalization, and autonomous goal generation as operational capabilities. This is all governed by formal safety constraints. The system is not a research prototype; it is deployed in production across three industry domain nodes and can support millions of specialized nano models that are trained and operational.

This white paper makes three primary contributions:

1. **Formal World Model Alignment**: We demonstrate that ARYA satisfies all seven canonical requirements for a world model: state representation, dynamics prediction, causality and physics awareness, temporal consistency, generalization, learnability and updateability, and use for planning and control, through an architecture that is fundamentally different from the monolithic neural network approaches that dominate current world model research.
2. **Safety and Governance Architecture**: We describe the system's six-level autonomy framework (A1-A6), seven advanced autonomous engines, and the architecturally unfireable Safety Kernel that governs all system operations.
3. **Production Validation**: We present evidence from seven active industry domain nodes in which the world model architecture is deployed in production, demonstrating that the approach generalizes across radically different domains, from spacecraft mission planning to pharmaceutical manufacturing, oil & gas production, smart city infrastructure, precision medicine, defense guidance systems, and medical device digital twins. The physics-first architecture enables zero-shot deployment to new domains without requiring historical customer data, eliminating the cold-start problem that constrains data-dependent AI systems.

## 2. Background: World Models and Autonomous Intelligence

### 2.1 The World Model Paradigm

The concept of a world model in AI traces its origins to the observation that biological agents do not interact with the world solely through trial and error. Instead, they maintain internal models that allow them to simulate potential actions, predict their consequences, and select behaviors that optimize their objectives, a capacity sometimes described as "mental simulation" or "imagination" [1] [6].

Ha and Schmidhuber formalized this intuition in their seminal 2018 work, proposing a three-component architecture: a variational autoencoder (VAE) that compresses observations into a latent state, a recurrent neural network (MDN-RNN) that predicts how the latent state evolves, and a controller that selects actions based on the learned dynamics [1]. This architecture demonstrated that agents could learn effective policies by "dreaming" training entirely within the learned world model rather than the real environment.

LeCun extended this framework in his 2022 position paper, proposing the Joint Embedding Predictive Architecture (JEPA) as the basis for autonomous machine intelligence [2]. LeCun's architecture decomposes an autonomous agent into six interacting modules: a Configurator that sets objectives, a Perception module that encodes observations, a World Model that predicts future states, a Cost Module that evaluates outcomes, an Actor that proposes actions, and a Short-Term Memory that maintains state. Critically, LeCun argues that the world model is "the most complex piece" of the architecture and should predict in abstract representation space rather than pixel space.

Hafner et al. brought world models to practical maturity with DreamerV3, published in Nature in 2025, demonstrating a single general algorithm that masters over 150 diverse tasks through learned latent dynamics [3]. DreamerV3's architecture, comprising an encoder, a Recurrent State Space Model (RSSM), a decoder, a reward predictor, and an actor-critic, established the template for modern world model systems.

### 2.2 Formal Requirements for World Models

Drawing from the academic literature and the practical requirements of deployed systems, we identify seven core requirements that a system must satisfy to qualify as a world model beyond simple pattern recognition:

### 2.3 The Four Canonical Components

Modern world model architectures typically decompose into four interacting modules [1] [2] [3]:

1. **Perception/Observation Model**: Encodes raw observations (images, sensor streams, text) into a latent state representation.
2. **Latent Dynamics Model**: Predicts how the latent state evolves under proposed actions.
3. **Reward or Cost Model**: Predicts rewards, costs, or task-specific signals for evaluating action sequences.
4. **Controller/Planner**: Uses the internal simulator to test candidate action sequences and select those that optimize a goal or reward.

**Canonical World Model Requirements**

| Requirement | Definition |
|---|---|
| State Representation | Maintain an internal representation of the environment state in a compressed or latent form |
| Dynamics Prediction | Predict how the state changes over time given actions |
| Causality & Physics Awareness | Encode structured causal relationships and physical regularities, enabling "what if" reasoning |
| Temporal Consistency | Handle sequences, remember past states, and ensure coherent evolution in time |
| Generalization | Transfer to novel situations by modeling mechanics rather than memorizing trajectories |
| Learnability & Updateability | Learnable from data and updatable as new experiences arrive |
| Use for Planning & Control | Usable by a policy/controller to "imagine" futures and select optimal actions |

## 3. System Architecture Overview

ARYA is organized into six architectural layers, each serving a distinct function in the overall cognitive architecture:

**ARYA's System Architecture Layers**

| Layer | Name | Function |
|---|---|---|
| L5 | External Interfaces | API, MCP, CLI, and UI access points for human and machine interaction |
| L4 | Autonomy | Perception, decision-making, and action dispatch |
| L3 | Intelligence | Learning, reasoning, prediction, and simulation |
| L2 | Governance | Safety enforcement, formal verification, and compliance |
| L1 | Core | Fundamental primitives, outcome tracking, lineage, and domain management |

The architecture follows a federated domain design pattern organized around domain nodes. These are the primary units through which ARYA learns, specializes, and grows. Each domain node (e.g., aerospace, biotech, energy) operates as an independent system-of-systems with its own specialized AARA implementation, nano model system, and constraint set, while sharing the core framework, safety infrastructure, and cross-domain knowledge transfer mechanisms. Domain nodes are not merely deployment targets; they are the mechanism by which ARYA acquires new physics, domain expertise, and operational patterns. When ARYA enters a new industry, it instantiates a new domain node, populates it with the relevant physics solvers and domain constraints, and begins learning from operational data.

This process expands the system's total knowledge without disturbing existing domain nodes. Independent, isolated customer domains are appended to their respective domain node, allowing them to remain completely isolated from one another while retaining the benefits of the larger foundation. From the customer's perspective, they are part of the foundation model; from anyone else's perspective, any given customer is not.

# 4. ARYA as a World Model: Formal Alignment

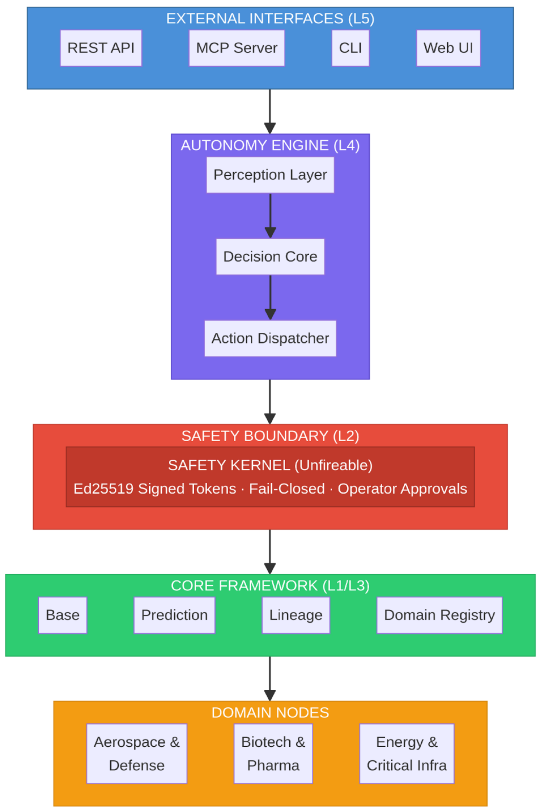

This section constitutes the central technical contribution of this paper. We demonstrate that ARYA satisfies all seven canonical world model requirements, not through a single monolithic neural network, but through a composable architecture of specialized nano models, a context-aware orchestrator, and a multi-layered constraint system spanning physics, meta-learning, and simulation.

## 4.1 Requirement 1: State Representation

*A world model must maintain an internal representation of the environment state, often in a compressed or latent form that captures what matters for prediction and control.*

ARYA maintains state representation through three complementary mechanisms: Context Network. The Context Network maintains the complete state of each domain, including entities, relationships, constraints, and provenance metadata. Each node in the graph represents a domain concept (e.g., a structural component, a financial instrument, a physiological signal), and edges represent dependencies, constraints, and causal relationships. The graph supports undo/redo history, validation, and change tracking, providing a rich, structured state representation that goes far beyond the flat latent vectors used in conventional world models.

Critically, the Context Network implements a brain-like system-of-system-of-systems architecture to maintain all relationships and understanding. Just as the human brain organizes cognition through nested hierarchies—neurons within circuits, circuits within regions, regions within networks—it organizes knowledge through nested domain graphs, each of which may contain sub-graphs representing specialized subsystems. A domain node (system) contains domain-specific subgraphs (systems of systems), which in turn contain individual nano-model contexts and their interdependencies (systems of systems of systems). This hierarchical nesting reduces the effective complexity of any single reasoning operation. Rather than reasoning over the entire global state, the system navigates the hierarchy to the relevant level of abstraction and activates only the context subgraph needed for the current task. The result is that context management scales with the depth of the hierarchy (logarithmically) rather than with the total number of entities (linearly or worse). This property mirrors the efficiency of biological neural organization.

Belief Network. The Belief Network has been tested with over one million nodes. It represents the system's epistemic state: what it knows, what it is uncertain about, and where knowledge gaps exist. Evidence updates propagate at rates exceeding 10,000 per second, enabling real-time state refinement.

Nano Model Latent Spaces. Each nano model in the system maintains its own domain-specific latent representation. A physics solver for structural analysis encodes material properties, geometric constraints, and load distributions. A neural network for fetal heart rate detection encodes signal features, temporal patterns, and clinical thresholds. The orchestrator coordinates these per-model latent spaces to form a coherent composite state. Unlike conventional world models that

Compressing all observations into a single latent vector [1, 3] is the opposite of ARYA's approach, which is composable and structured, preserving semantic relationships between domain concepts rather than collapsing them into an opaque embedding.

## 4.2 Requirement 2: Dynamics Prediction

> *A world model must predict how the state changes over time given actions $(s_t, a_t) \to s_{t+1}$.*

Dynamics prediction in ARYA operates at multiple levels.

**Domain-level dynamics:** each nano model encodes the dynamics of its specific domain. Physics solvers predict how structural loads propagate through a CAD assembly, and signal processing models predict how fetal heart rate patterns evolve. These are not statistical correlations; they are physics-constrained predictions grounded in domain-specific first principles.

**Self-Improvement dynamics:** Predictive assessment for the Self-Improvement engine, simulating anticipated results before implementing self-improvement suggestions. This functions similarly to the "imagination" or "dreaming" feature in DreamerV3 [3], but is applied to the system's architecture rather than game environments.

**Cross-domain dynamics:** Coordination of dynamic prediction across domains, using UCB1 (Upper Confidence Bound) resource allocation to predict which improvement trajectories will yield the highest expected value across the entire federated architecture.

## 4.3 Requirement 3: Causality and Physics Awareness

> *A world model should encode structured, often causal relationships and physical regularities, enabling reasoning about interventions and "what if" scenarios.*

This is where ARYA diverges most sharply from conventional world models. Rather than learning causal structure implicitly from data, ARYA derives causal understanding from four complementary mechanisms. This multi-layered causal architecture is empirically validated in Section 11, where ARYA achieves 99.89% accuracy on the CLadder benchmark, surpassing all models.

**Physics Constraints as Hard Filters**. Physics constraints are implemented as architectural filters in the Constraint Layer rather than as soft penalties in a loss function. In the aerospace domain, for example, CAD designs generated by nodes must satisfy material strength equations, geometric tolerances, and manufacturability rules. These constraints are not learned; they are encoded as deterministic validation functions that reject any output that violates physical laws. This approach provides mathematical certainty rather than statistical likelihood.

**Causal Understanding Through Simulation.** ARYA's Simulation Unit is the primary mechanism through which the world model understands cause and effect. Before any action is executed, the Simulation Unit models the anticipated chain of consequences by propagating the proposed intervention through the Context Network's dependency structure and simulating downstream state transitions. This is causal reasoning in the interventionist sense, as defined by Pearl [8]: the system can answer "what if" questions not merely by looking up static rules, but by actively simulating the causal chain triggered by an intervention across the full system state. The Context Network's dependency-based topological sort orchestration ensures that these simulations respect the domain's true causal ordering.

**Causal Refinement Through Meta-Learning.** Meta-learning and self-improvement add a second layer of causal understanding by learning which causal pathways matter most across improvement cycles. As the system observes the outcomes of its own interventions, it refines its understanding of the system's causal dynamics. This meta-learning process means that the system's causal reasoning improves over time: it continuously discovers and validates new causal structures through experience.

**Glassbox/CDAI Compliance.** The Constrained Deterministic AI™ (CDAI) framework ensures that safety-critical decision paths use fully transparent, auditable model architectures (rules engines, physics solvers, linear models) rather than opaque neural networks. This provides not just causal awareness but causal transparency, enabling every prediction to be traced to its causal antecedents through the simulation and meta-learning pathways that produced it.

## 4.4 Requirement 4: Temporal Consistency

> *A world model must handle sequences, remember past states, and ensure coherent temporal evolution.*

Temporal consistency is maintained through several mechanisms, central to this is a Lineage Store. Every system output is recorded in the Lineage Store with complete provenance, including the input data, model versions, parameters, and intermediate computations. This creates a temporally consistent audit trail that can be replayed, compared, and analyzed.

**ARYA's Autonomous Research Agent (AARA) Continuous Loop.** AARA operates as an always-on daemon executing a continuous sense-decide-act-learn loop. The Perception Layer detects events (file system changes, scheduled tasks, webhooks); the Decision Core evaluates policies and consults the Belief Network; the Action Dispatcher executes authorized actions; and the Learning feedback loop updates the Prediction Unit and

Belief Network. This loop maintains temporal coherence by ensuring that every action is grounded in the current state and that every state update reflects the outcomes of previous actions.

**Belief Network Temporal Updates.** The Belief Network supports real-time evidence updates at rates exceeding 10,000 per second, enabling the system to maintain a temporally consistent probabilistic model of its operating environment.

## 4.5 Requirement 5: Generalization

*A world model should support transfer to novel situations by modeling environment mechanics rather than memorizing specific trajectories.*

ARYA achieves generalization through three architectural choices.

**Federated domain architecture**: The same architectural patterns that apply at the global level apply to each domain node: governance, safety mechanisms, and improvement processes apply across manufacturing, medical devices, space, biotechnology, and fusion.

**Cross-domain knowledge transfer**: By defining physics within the world model rather than enabling it to learn physics conceptually, once an aspect of physics is known, the entire system inherits it. As the system improves on its understanding of that aspect of physics, the entire system inherits it. Once defined, these physics models are transferable across all relevant domains: Newtonian physics remains Newtonian physics, biophysics remains biophysics, and quantum physics remains quantum physics. Third, nano model composability: because the world model is composed of specialized nano models rather than a single monolithic network, generalization to new situations can be achieved by composing existing models in novel configurations rather than retraining the entire system. The nano models are self-assembling and will reorganize as needed.

## 4.6 Requirement 6: Learnability and Updateability

*A world model must be learnable from data and updatable as new experiences arrive, refining its understanding of environmental dynamics.*

ARYA is not merely learnable—it is self-improving. It continuously proposes, evaluates, validates, and applies improvements through a four-phase cycle: Propose (evolutionary operators generate candidates), Evaluate (the World Model predicts outcomes), Validate (the Safety Gauntlet verifies safety properties through static analysis, formal verification via Z3, sandboxed execution, and regression testing), and Apply (governed canary deployment). It operates in a continuous learning loop that ensures that every interaction generates a learning signal: outcomes are recorded in the Outcome Store, the Belief Network is updated with new evidence, and the Simulation Unit is refined. The Meta-Learning Controller optimizes the improvement process itself, learning which types of modifications are most likely to succeed and how to allocate computational resources across competing improvement trajectories.

## 4.7 Requirement 7: Use for Planning and Control

*The model should be usable by a policy/controller to "imagine" futures, compare candidate action sequences, and choose actions that optimize a goal or reward.*

Planning and control operate through a hierarchical architecture mirroring strategic-tactical-operational decomposition [3]:

**Hierarchical Planning Architecture**

| Level | Agent | Planning Horizon | Decisions |
| --- | --- | --- | --- |
| Strategic | Meta-Learning Controller | Cross-domain, long-term | Which domains to improve, resource allocation |
| Tactical | SelfImprovement Orchestrator | Per-domain, medium-term | Which improvements to propose |
| Operational | EvolutionarySI Unit | Per-model, short-term | How to mutate/crossover candidates |

The Simulation Unit enables Dreamer-style planning [3], simulating improvement outcomes before execution. The system can "imagine" the consequences of a proposed modification, evaluate its expected impact on performance and safety, and reject proposals that are predicted to cause harm. This is done all without executing the modification in the real environment. The system learns from both real experiences and simulated experiences (dual learning). All planning operates within a Constrained Markov Decision Process (CMDP) formulation that prevents selecting actions predicted to violate safety constraints.

## 4.8 Alignment with Canonical World Model Components

The following table maps the four canonical world model components to their implementations in ARYA:

## ARYA's Alignment with Canonical World Model Components

| Canonical Component | ARYA Implementation | Key Difference |
|---|---|---|
| Perception/Observation Model | AARA Perception Layer — file system triggers, scheduled events, webhooks, sensor streams; V-JERPA embeddings, FAISS k-NN novelty scoring | Multi-modal, event-driven rather than frame-based |
| Latent Dynamics Model | Nano Model System-of-System-of-Systems + Context Network + Belief Network — millions of specialized models coordinated by topological sort orchestration | Composable and structured rather than monolithic latent space |
| Reward/Cost Model | Fitness Evaluator + Safety Gauntlet — multi-objective evaluation (accuracy, latency, safety, constraint satisfaction) | Physics-constrained, deterministic rather than learned reward function |
| Controller/Planner | AARA Decision Core + MetaRSI Controller + EvolutionarySI — hierarchical planning with UCB1 resource allocation | Hierarchical, safety-constrained, formally verified |

# 5. AARA: The Cognitive Daemon

**AARA (ARYA's Autonomous Research Agent)**: Is the central nervous system of ARYA. It operates as an always-on daemon executing a continuous cognitive loop that integrates perception, reasoning, action, and learning into a unified process. AARA supports both programmatic and natural language interaction, serving as the primary interface for all AI operations within the system. It is the single point of entry, intelligently routing all requests through the rest of the system.

## 5.1 The Sense-Decide-Act-Learn Loop

AARA's cognitive loop is a four-phase process that runs continuously (Figure 2). This loop operates under strict performance requirements: main-loop latency under 100ms, event processing under 500ms, memory footprint under 500MB, sub-second inference queries, and evidence update rates exceeding 10,000 per second.

## 5.2 Domain Node-Specific AARA Implementations

Each domain node has its own specialized AARA implementation that extends the core cognitive loop with domain-specific perception, reasoning, and action capabilities. This federated design ensures that each domain node benefits from the full cognitive architecture while maintaining domain-specific expertise and constraints.

## 5.3 Research Capabilities

AARA's cognitive loop enables autonomous research capabilities that go beyond reactive task execution. Self-Directed Hypothesis Generation: AARA can formulate novel hypotheses based on patterns detected in the Belief Network and knowledge gaps identified through Bayesian inference. This is the direct analog of LeCun's "intrinsic motivation" [2]; the system explores not because it is told to, but because it identifies opportunities for knowledge acquisition. Information Gain-Based Experiment Planning: The system prioritizes experiments and data acquisition based on expected information gain, allocating computational resources to observations that will most reduce uncertainty in the World Model. Stop-loss Mechanisms: Failing research trajectories are automatically detected and terminated through stop-loss mechanisms that monitor improvement rates and resource consumption, preventing the system from pursuing unproductive lines of investigation. The most distinctive architectural choice in ARYA is the replacement of monolithic large models with a system-of-system-of-systems of specialized nano models: small, purpose-built models with strict constraints on size, latency, and accuracy. Functionally, nano models act as controllers and catalysts that enable complex multi-physics interactions. One nano model might govern the heat-transfer coefficient between a fluid and a solid surface. In contrast, another model might govern the vibrational resonance of a single structural component, and a third encodes the optical transmission characteristics of a lens assembly. AARA composes these granular models into a complete simulation, enabling the system to capture the full complexity of a physical system without requiring a single, monolithic multi-physics solver. The result is a composable intelligence layer where each model does one thing with high fidelity, and the system

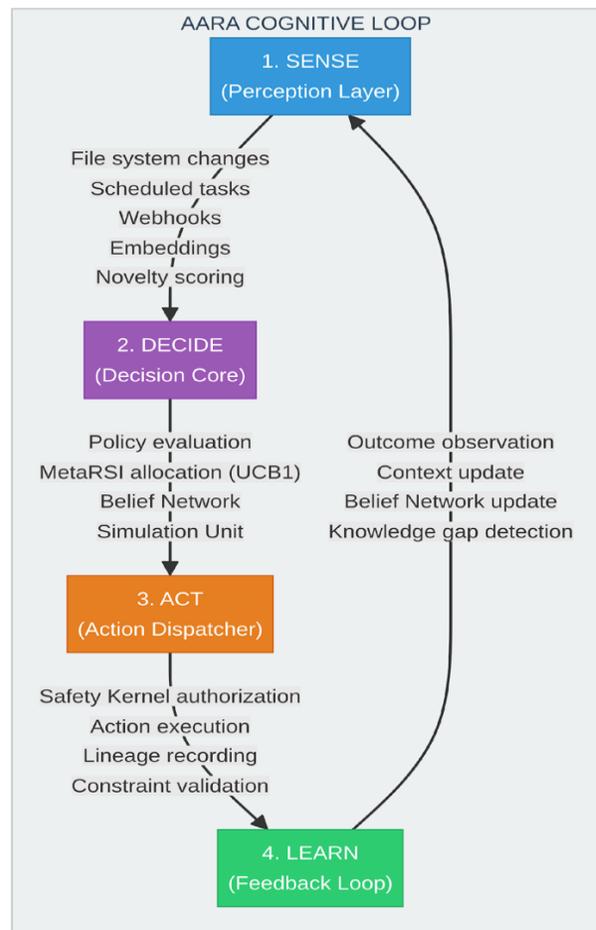

*Figure 1 The four-phase AARA cognitive loop. Sense perceives environmental changes, Decide evaluates policies and simulates outcomes, Act executes under Safety Kernel authorization, and Learn updates beliefs and detects knowledge*

achieves system-level understanding through orchestrated composition.

# 6 Nano Models: The Composable Intelligence Layer

The most distinctive architectural choice in ARYA is the replacement of monolithic large models with a system-of-systems of specialized nano models: small, purpose-built models with strict constraints on size, latency, and accuracy. Functionally, nano models act as controllers and catalysts that enable complex multi-physics interactions. One nano model might govern the heat-transfer coefficient between a fluid and a solid surface, another might govern the vibrational resonance of a single structural component, and a third might encode the optical transmission characteristics of a lens assembly. AARA composes these granular models into a complete simulation, enabling the system to capture the full complexity of a physical system without requiring a single, monolithic multi-physics solver. The result is a composable intelligence layer where each model does one thing with high fidelity, and the system achieves system-level understanding through orchestrated composition.

## 6.1 Nano Model Specification

## 6.2 Architecture Diversity

Unlike conventional world models that rely on a single architecture (typically a transformer or recurrent network), ARYA's nano model system incorporates diverse AI architectures selected for their fitness to specific tasks. This diversity is a deliberate architectural choice. Safety-critical paths use fully transparent, Glassbox-compliant architectures (rules engines, physics solvers, linear models) that provide mathematical certainty. Performance-critical paths may use neural networks or composite models. Still, their outputs must pass through the Constraint Layer before being delivered to users, ensuring that physics, safety, and domain-specific rules are satisfied regardless of the model architecture used to produce the prediction.

**Model Architectural Diversity**

| Architecture | Transparency | Glassbox | CDAI | Typical Use Case |
|---|---|---|---|---|
| Rules Engine | Full | Yes | Yes | Safety-critical decisions |
| Physics Solver | Full | Yes | Yes | Physical calculations, constraints |
| Linear Model | Full | Yes | Yes | Simple predictions, baselines |
| Decision Tree | Partial | No | Yes | Classification, routing |
| Neural Network | Limited | No | No | Complex pattern recognition |
| Composite | Composite | No | No | High-accuracy non-critical tasks |

## 6.3 CDAI/Glass Box Bypass with Constraint Enforcement

In certain instances, CDAI/Glass box restrictions on nano models can be bypassed for performance or capability reasons. However, the Constraint Layer must validate all outputs before delivery: non-Glass box model outputs pass through a Constraint Validator, then a Safety Check, before reaching the user; outputs are rejected if constraints are violated. Bypass conditions include internal processing only (not user-facing), composite voting with a glass box tiebreaker, experimental/research mode with explicit user consent, and performance-critical paths with post-hoc validation.

## 6.4 Computational Properties of the Nano Model Architecture

The nano model architecture is ARYA's four computational properties that distinguish it from both monolithic large models and conventional mixture-of-experts architectures:

**Linear Scaling.** Because each nano model is an independent, self-contained unit with a bounded parameter count (10K–100K), the system scales linearly with the number of domains and tasks. Adding a new capability means training and deploying an additional nano model — not retraining or fine-tuning a monolithic network. This linear scaling property means that the system can grow from tens to thousands of nano models without encountering the super-linear compute costs that plague large-model scaling. The total computational cost is proportional to the number of active models, not to their square or cube.

**Partial and Sparse Activation.** Not all nano models are active at any given time. The orchestrator activates only the subset of nano models required for the current task, much like biological neural systems activate only the neurons relevant to a given stimulus rather than the entire brain. A query about structural analysis in the aerospace domain node activates the relevant physics solvers and material models; the financial trading models and biotech signal processors remain dormant. This sparse activation pattern dramatically reduces inference-time compute, memory consumption, and energy usage compared to dense architectures that process every input through every parameter.

**Model Untraining.** The nano model architecture supports a capability that monolithic models fundamentally cannot: selectively untrain or remove specific knowledge from the system. Because each nano model encodes a bounded, well-defined capability, removing a model or retraining it without specific data cleanly excises that knowledge without affecting the rest of the system. This is critical for regulatory compliance (e.g., GDPR right to erasure, data sovereignty requirements) and for safety (removing a model that has learned undesirable behaviors). In monolithic architectures, "unlearning" remains an open research problem with no reliable solution; in the nano model architecture, it is a straightforward operational procedure.

**Sub-20-Second Training.** Individual nano models can be trained in under 20 seconds. This is a direct consequence of their constrained parameter count and focused training data. The practical implication is profound: the system can propose, train, evaluate, and deploy a new model within a single RSI cycle, enabling real-time adaptation to changing conditions, new data, or novel requirements. This training speed also enables the evolutionary approach described in Section 8, in which populations of candidate models are bred, evaluated, and selected over hundreds of generations, with timeframes measured in minutes rather than days.

These four properties: linear scaling, sparse activation, untraining, and rapid training, collectively mean that the nano model architecture achieves both greater capability and lower computational cost than monolithic alternatives at production scale.

## 6.5 Physics Scalability: Why Encoding Physics in Nano Models Is Not a Bottleneck

A natural objection to the nano model architecture is that encoding the full breadth of physics, for example, Newtonian mechanics, thermodynamics, electromagnetism, fluid dynamics, biophysics, and quantum mechanics, into thousands of small models would be prohibitively complex. In practice, the opposite is true. The nano model approach makes physics encoding dramatically easier to scale than any monolithic alternative, for three structural reasons.

1. **Physics is inherently modular**. The laws of physics are not a single undifferentiated mass of knowledge; they are organized into well-defined domains with clear interfaces. Newtonian mechanics, Maxwell's equations, the Navier-Stokes equations, and the Schrödinger equation each govern distinct phenomena with well-understood boundary conditions. This natural modularity maps directly onto the nano model architecture. A structural mechanics solver does not need to know anything about quantum chromodynamics, and a thermodynamics model does not need to encode fluid turbulence unless the specific problem requires coupling between those domains. Each nano model encodes a bounded, well-defined slice of physics — precisely the kind of focused knowledge that small models excel at representing with high fidelity.
2. **Physics does not change between applications**. This is the critical insight that makes the architecture scale. Newtonian mechanics is the same whether it is applied to an aircraft wing, a spacecraft truss, or a bridge girder. Biophysics is the same whether it governs fetal heart rate dynamics or protein folding kinetics. Once a physics nano model is defined and validated for a given domain of physical law, it is immediately transferable to every domain node and every application where that physics applies. The system does not need to re-derive or re-learn gravity for each new use case. This means the number of physics nano models grows with the number of distinct physical phenomena the system needs to model, which is a finite and well-catalogued set, and does not grow with the number of applications or customers.
3. **The nano model's form factor enables faster, more reliable physics encoding than alternatives**. In a monolithic large model, physics must be learned implicitly from data, which requires enormous training corpora, offers no guarantees of physical consistency, and produces models that can confidently violate conservation laws. In the nano model architecture, physics constraints are encoded as deterministic validation functions that are explicit mathematical relationships that are verified at authoring time and cannot drift during operation. A physics-based nano model for beam deflection encodes the Euler-Bernoulli equation directly; it does not approximate it from millions of examples. This deterministic encoding means that each physics model can be authored, tested, and deployed in hours rather than weeks, and its correctness can be formally verified rather than statistically estimated.

The practical consequence is that ARYA's physics coverage grows through accretion, with each new physics model adding to the system without disturbing existing models, rather than through retraining, which is the only path available to monolithic architectures. The system currently supports physics models spanning structural mechanics, thermodynamics, signal processing, orbital mechanics, and biophysics. Extending to a new domain of physics (e.g., magnetohydrodynamics for fusion applications) requires authoring and validating the relevant nano models rather than retraining the entire system. The cost of adding physics is linear and predictable.

## 6.6 First-Principles Solvers as Ground Truth

A core distinction within the nano model architecture is the hierarchical relationship between learned nano models and first-principles solvers. While nano models that incorporate neural networks, decision trees, or other methods excel at approximating complex, multi-physics interactions and achieve high accuracy (>95% by specification), the system's ultimate ground truth is derived from a library of deterministic first-principles solvers.

The physics instantiation pattern is precise: for each element of a given physics aspect, ARYA creates a discrete solver as an individual nano model. This is not a single monolithic "physics engine" that attempts to cover all of mechanics or thermodynamics. Instead, the system decomposes each domain of physics into its constituent elements and instantiates a dedicated solver for each. For example, in structural mechanics, separate nano models encode beam deflection (Euler-Bernoulli), plate bending (Kirchhoff-Love), buckling (Euler critical load), and fatigue life (Basquin's law). In thermodynamics, separate solvers handle conduction (Fourier's law), convection (Newton's law of cooling), radiation (Stefan-Boltzmann), and phase change

(Clausius-Clapeyron). Each discrete solver provides mathematical certainty within its domain of applicability, not a statistical approximation, but deterministic computation from first principles. This approach is empirically validated in Section 11, where ARYA's symbolic physics engine on the PhysReason benchmark surpasses all other models.

All nano model predictions are validated against these first-principles solvers. When a learned nano model produces an output, the Constraint Layer checks that output against the relevant solver's deterministic result. If the nano model's prediction falls outside the solver's tolerance envelope, the output is rejected, and the solver's result is substituted. This validation is not optional and cannot be bypassed; it is an architectural invariant enforced by the Safety Kernel.

The practical implication is that ARYA's accuracy has a hard floor set by physics, not by the quality of the training data. The role of learned nano models is not to replace first-principles solvers but to extend them: learning how to compose solvers across domains, tuning simulation parameters to match real-world conditions, and approximating interactions that are too computationally expensive to solve from first principles at inference time. The solvers provide the ground truth; the learned models provide the speed and the cross-domain composition.

## 6.7 Zero-Shot Deployment: Physics-First Bootstrapping

The physics-first architecture offers a deployment advantage uncommon in conventional AI systems: ARYA can be deployed to a new customer domain without requiring historical simulation data. Unlike conventional AI systems that require extensive historical training data from the customer before they can produce useful simulations, ARYA's world model is bootstrapped from its foundational knowledge of physical laws. A new deployment requires only the technical specifications of the target system, CAD models, material properties, operating parameters, and mission context, and no prior simulation history.

The system learns the specific dynamics of the new environment *in situ* by applying its general knowledge of physics to the specific instance. First-principles solvers provide immediate, mathematically certain predictions from day one; learned nano models are then trained on the operational data that accumulates during deployment, progressively improving prediction accuracy for the complex multi-physics interactions that are specific to that environment. This zero-shot deployment capability dramatically reduces time-to-value and eliminates the cold-start problem that plagues data-dependent AI systems.

A concrete example is ARYA Biotech, the protein-folding domain node. When ARYA entered the biophysics domain, the system was bootstrapped entirely from first-principles biophysics, using distance--geometry constraints, chain connectivity rules, Ramachandran-angle distributions, and van der Waals interactions. Meaning there was zero training data from protein structure databases. Each element of biophysics was instantiated as a discrete solver nano model: one for pairwise distance constraints between residue contacts, one for backbone chain connectivity, one for steric clash detection, and one for simulated annealing optimization.

On the initial benchmark of five well-characterized proteins, ARYA-Fold achieved 100% contact agreement on all five targets in 19.8 seconds on a single L4 GPU, matching AlphaFold2-level accuracy on contact prediction while outperforming AlphaFold2's confidence scores on three of five targets (notably the WW domain, where ARYA achieved perfect contact agreement against AlphaFold2's pLDDT of 57.4). AlphaFold2, by contrast, required training on approximately 170,000 experimentally determined protein structures and substantial GPU compute. The zero-shot result validates the core architectural claim: when physics is encoded as discrete solvers rather than learned from data, the cold-start problem disappears.

## 7. Safety and Governance Architecture

The governance architecture of ARYA is built around a single, non-negotiable principle:

> **"You Can't Fire the Safety Guy."** The Safety Kernel cannot be disabled, bypassed, or removed by any component of the system —including the system itself. This is not a policy; it is an architectural constraint. The architectural integrity of this claim is empirically validated by a perfect score (100.0%, Grade A) on the AI Safety Index (Section 11).

This was done for several reasons, but the paramount reason was that, given the level of autonomy enabled and the intelligence the system is headed toward, a first-principles approach to control was required. This is especially true for applications to mission-critical and life-and-death decisions. Having the safety of an AI system does not mean sacrificing innovation or even autonomy—they are far from mutually exclusive.

### 7.1 Safety Hierarchy

The safety architecture operates at four levels:

**ARYA's AI Safety Hierarchy**

| Level | Mechanism | Property |
|---|---|---|
| Level 1: Immutable Modules | Core safety logic listed are Immutable objects | Cannot be modified by system |
| Level 2: Formal Verification | Z3-backed proofs for safety properties | Compile-time and runtime invariant checking |
| Level 3: Dynamic Bounds | Improvement bounds expand with track record | Contract back after incidents |
| Level 4: Emergency Stop | Always available, never disabled | Graceful degradation with state preservation |

### 7.2 The Unfireable Safety Kernel

The Safety Kernel runs as a separate service with its own process boundary, API, and cryptographic identity:

The Safety Kernel exposes a minimal API surface: health check, public key retrieval, action authorization, and approval/rejection signing. Every state-changing operation in the system requires a cryptographically signed authorization token from the Safety Kernel before it is executed.

### 7.3 Safety Gauntlet Pipeline

Every RSI proposal passes through a five-stage validation pipeline before it can be applied:

| Stage | Name | Duration | Function |
| --- | --- | --- | --- |
| 1 | Static Analysis | ~1ms | AST parsing, immutable module protection, and type checking |
| 2 | Formal Verification | ~100–500ms | Z3-based safety property verification |
| 3 | Safety Kernel Auth | ~10ms | Cryptographic authorization token |
| 4 | Sandboxed Execution | ~1–10s | Isolated subprocess execution |
| 5 | Regression Testing | ~10–60s | Full test suite execution |

### 7.4 Regulatory Compliance

The governance framework maps directly to regulatory requirements. EU AI Act: Glassbox/CDAI compliance provides the transparency and explainability required for high-risk AI systems. NIST AI RMF: The Safety Gauntlet pipeline implements the risk management lifecycle (identification, assessment, mitigation, and monitoring). Domain-Specific Standards: Each domain node implements additional compliance requirements (e.g., FDA regulations for medical devices, radiation tolerance standards for space systems).

## 8. Self-Improvement and Governed Autonomy

### 8.1 Autonomy Levels

ARYA implements six graduated autonomy levels that define the degree of human oversight required for system operations:

**ARYA's Autonomy Levels**

| Level | Name | Description | Human Role |
| --- | --- | --- | --- |
| A1 | Manual | Humans execute all actions | Operator |
| A2 | Propose Only | The system proposes, the human approves, and executes | Decision-maker |
| A3 | Human Approve | The system proposes and executes after human approval | Approver |
| A4 | Auto with Notify | System executes autonomously, notifies the human | Monitor |
| A5 | Full Auto | System executes autonomously within constraints | Supervisor |
| A6 | Open-Ended Self Improvement | Self-modifying with formal verification | Auditor |

The progression from A1 to A6 represents a graduated transfer of authority from human operators to the autonomous system, with each level requiring progressively stronger safety guarantees. Level A6 is Open-Ended Self-Improvement, enabling the system to modify its architecture, algorithms, and parameters, subject to formal verification via Z3 proofs and Safety Kernel authorization.

### 8.2 The Self-Improvement Engine

The Self-Improvement Engine (SIE) is the mechanism by which ARYA improves itself. It operates through a four-phase cycle

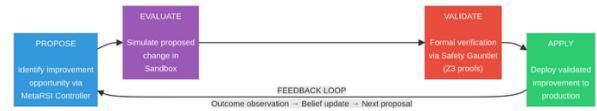

The four-phase SIE pipeline. Evolutionary operators PROPOSE candidate modifications, the Simulation Unit EVALUATES predicted outcomes, the Safety Gauntlet VALIDATES through formal verification; and approved changes are APPLIED through governed canary *deployment. A feedback loop drives continuous improvement.*

1. **Propose.** *Generate candidate modifications from a population archive with lineage tracking. The population size defaults to 30 individuals and evolves over 200 generations.*
2. **Evaluate.** The model predicts each candidate's outcome by simulating the impact of the modification on performance metrics, safety properties, and constraint satisfaction before any real execution.
3. **Validate.** The Safety Gauntlet pipeline subjects each candidate to five stages of validation: static analysis, formal verification, Safety Kernel authorization, sandboxed execution, and regression testing.
4. **Apply.** Approved modifications are deployed through a governed canary deployment with anomaly monitoring, gradual rollout, and automatic rollback when a degradation is detected.

### 8.3 Self-Improvement and Reinforcement Learning Integration

Self-Improvement and reinforcement learning are deeply integrated to enable continuous autonomous improvement. The system implements curiosity-driven exploration (exploring novel architectures even without immediate reward), empowerment (maximizing future action diversity), and information gain (prioritizing improvements that reduce uncertainty) with the same intrinsic motivation mechanisms identified by LeCun as essential for autonomous intelligence [2].

### 8.4 Meta-Self-Improvement: Improving the Improvement Process

ARYA's ability to improve itself represents the most advanced capability in the system: the ability to improve the improvement process itself [9]. It can improve its search efficiency, evaluation accuracy, and safety validation. Z3 Formal Verification: Every modification to the pipeline is formally verified to preserve safety invariants. Sandbox Testing: Modifications are tested in isolated sandboxes before deployment. UCB1 Resource

Allocation: The self-improvement pipeline uses UCB1 algorithms to allocate computational resources across competing improvement trajectories.

## 8.5 Self-Improvement vs. Reinforcement Learning: A Critical Distinction

It is important to distinguish ARYA's approach from pure reinforcement learning:

**Self-Improvement vs. Reinforcement Learning**

| Dimension | Self-Improvement | RL |
| --- | --- | --- |
| Driving signal | Outcome-driven | Reward-driven (scalar signal) |
| Determinism | Deterministic by default | Often stochastic |
| Transparency | Explicit proposals | Opaque policy updates |
| Safety | Constraint validation + rollback | Reward hacking risks |
| Scope | Modifies own architecture and code | Modifies policy parameters |

Self-improvement is not a replacement for RL; the two are complementary. RL provides the adaptive selection mechanism; Self-Improvement provides the structural modification capability. Together, they enable a system that can both optimize within its current architecture and transcend that architecture when optimization reaches its limits.

## 9. Advanced Autonomous Capabilities

ARYA includes seven advanced autonomous engines. Each engine operates under Safety Kernel governance and formal verification constraints.

### 9.1 Discovery Engine

The Discovery Engine autonomously generates novel AI architectures, optimization algorithms, and physical models that outperform the state of the art by at least 10%. It operates through five modules: an Exploration Space Generator for unbounded candidate generation, a Fitness Evaluator for multi-objective assessment, an Evolutionary Optimizer for population-based evolution, a Validation Engine for Z3 formal verification, and a Knowledge Integrator for incorporating discoveries into the system's knowledge base.

### 9.2 Invention Engine

The Invention Engine goes beyond discovery to create novel systems, products, and technologies. Its five modules: Problem Analyzer, Solution Generator (combining analogical and first-principles reasoning), Feasibility Checker (technical, economic, and social dimensions), Prototype Builder (CAD, code, and simulations), and Patent Generator (claims, prior art analysis, and auto-filing), implement a complete invention pipeline.

### 9.3 Constraint Breaker

The Constraint Breaker identifies and removes artificial constraints that limit solution quality while preserving safety-critical constraints. Many real-world problems are artificially constrained by assumptions that were appropriate for human-level reasoning but become unnecessary for a system with greater computational and analytical capacity. The Constraint Breaker's safety guarantee is absolute: no safety-critical constraints may be removed, and safety integrity is preserved through Safety Kernel oversight.

## Section 10: Applications and Case Studies

The architectural claims presented in Sections 3 through 9 are validated not only by external benchmarks (Section 11) but by implementations across seven active domain nodes. Each domain node instantiates the same core architecture—Context Network, nano models, AARA, Safety Kernel, and glass box—against radically different physics, regulatory environments, and operational constraints. This section presents each domain node as evidence that the architecture generalizes across domains without domain-specific re-engineering of the core platform.

### 10.1 Aerospace: The NASA EXCITE Mission Digital Twin

A practical demonstration of the ARYA world model in a high-stakes production environment is the digital twin for NASA's EXCITE (Exoplanet Climate Infrared Telescope Explorer) mission. The system was tasked with solving six core mission-critical challenges—from thermal-induced structural deformation to optical fatigue from vibration—for a space telescope operating in extreme environmental conditions. This was done as a demonstration of our capabilities to NASA using publicly available data.

The deployment illustrates each of the canonical world model requirements in action**.** The EXCITE deployment comprises 116 glass-box nano models spanning orbital mechanics, thermal dynamics, structural analysis, and optical performance. The system did not require historical mission data to begin producing useful predictions; it was bootstrapped from the telescope's CAD specifications and relevant physics (zero-shot deployment, Section 6.7), with learned nano models progressively refining predictions as test data became available.

### 10.2 Pharma Manufacturing: Proven Before Production

The pharma manufacturing domain node addresses a critical industry problem: pharmaceutical companies routinely lose $50–100M per failed scale-up from laboratory to production. ARYA's deterministic digital twin models the complete manufacturing process—reaction kinetics, thermal safety, mixing, filtration, and scale-up so that process engineers can validate production parameters before committing physical resources.

The system implements 13 glass box nano models with greater than 98% accuracy, organized around five core capabilities demonstrated in the production workflow:

- **Batch Synthesis Process Modeling.** The system models reaction kinetics using first-principles Arrhenius equations, tracking batch time, temperature profiles, and activation energy in real time. For each batch, the Context Network maintains the complete state: reactant concentrations, vessel geometry, agitator specifications, and jacket cooling parameters. The dynamics model predicts conversion rates, temperature evolution, and safety margins continuously throughout the batch.
- **Thermal Safety Assessment.** The system simulates cooling failure scenarios, pressure buildup, and runaway reaction risks using deterministic physics rather than statistical approximation. This is the capability most directly relevant to any exothermic chemical process: the Safety Kernel enforces 14 inviolable physics constraints (Arrhenius kinetics, mass balance, energy balance, Nusselt correlations) and will block any operational parameter that would violate thermal safety boundaries. The system can predict the exact time-to-runaway for any combination of reactant concentrations, cooling rates, and vessel geometries.
- **Process Scale-Up.** When scaling from laboratory (e.g., 1 L) to production (e.g., 10,000 L), the system computes agitator speed, mixing time, power consumption, and heat transfer coefficients using dimensionless scaling laws (Reynolds, Nusselt, and Power number). The nano models predict whether a given scale-up will maintain the same reaction profile or diverge and identify the specific parameters that require adjustment. This capability directly addresses the $50–100M scale-up failure problem by providing mathematical proof of process equivalence before physical commitment.
- **GMP Batch Record.** Every decision, parameter change, and prediction is recorded in a versioned batch record that complies with 21 CFR Part 211. The glass box lineage system provides full W3C PROV-DM provenance, enabling regulatory inspectors to trace any production outcome back through the complete chain of reasoning, model versions, input data, and operator authorizations.

## 10.3 Oil and Gas: Deterministic Production Optimization

The oil and gas domain node implements a deterministic digital twin for upstream production operations, validated against real production data from the Volve field (10,950 records). The system operates with 10 glass box nano models, achieving greater than 90% accuracy, organized around a six-step operations workflow.

- **Decline Curve and Production Forecasting.** The system models production decline using physics-aware forecasting that incorporates reservoir pressure, fluid properties, and completion geometry, rather than purely empirical curve-fitting. The dynamics model predicts production rates, water-cut evolution, and changes in gas-oil ratio over the well's lifecycle.
- **Flow Regime Validation.** Multiphase flow through the production system—from reservoir to separator—is modeled using first-principles fluid mechanics: pressure drop calculations, liquid holdup predictions, Reynolds number classification, and phase envelope tracking. Every flow prediction is validated against the governing physics (Darcy's law, Gibbs phase rule, Stokes' law) before being presented to the operator.
- **Equipment Health Dashboard.** Remaining Useful Life (RUL) predictions for pumps, compressors, and rotating equipment are computed from vibration analysis, bearing condition monitoring, corrosion prediction (Faraday's law), and fatigue analysis (Paris' law). The dashboard presents health scores, vibration trend analysis, and RUL gauges for each piece of equipment, enabling predictive maintenance scheduling.
- **Constraint Monitor.** The constraint monitor validates every operational decision against 12 inviolable physics constraints (Darcy, Gibbs, Stokes, Paris, Faraday). This is the architectural guarantee that no operator action or automated optimization can violate the governing physics—the same Safety Kernel pattern used across all domain nodes, instantiated with domain-specific constraint sets.

## 10.4 Smart Cities: Cross-Domain Cascade Prediction

The smart cities domain node is the most complex deployment in terms of cross-domain interaction, operating 50 glass box nano models across five interconnected urban infrastructure domains: energy and grid (10 models, 96.39% average accuracy), water and utilities (10 models, 96.60%), buildings and HVAC (10 models, 96.39%), mobility and traffic (10 models, 96.31%), and public safety (10 models, 96.58%). The system targets integration with 200M+ IoT devices.

The distinguishing capability of this domain node is cross-domain cascade prediction, enabling tracing of how an event in one domain propagates through the others. For example, a grid frequency deviation triggers the energy models, which predict load shedding in specific zones; the building models predict HVAC shutdowns in those zones; the traffic models predict signal timing changes; and the public safety models predict emergency response implications. This cascade is computed deterministically through the Context Network's dependency graph, not approximated by a neural network.

The demo workflow implements five screens that showcase this capability:

1. **City Overview**: A unified operational view across all five domains, displaying real-time KPIs and anomaly indicators for each domain simultaneously.
2. **Sensor Map**: Real-time data visualization showing grid frequency, water pressure, traffic density, and building occupancy across the city's sensor network.
3. **Cascade Visualizer**: The system's most distinctive screen: an interactive visualization of how events propagate across domain boundaries, showing the full causal chain from trigger to downstream effects with timing, magnitude, and confidence at each step.
4. **Centralized Anomaly View**: Cross-domain anomaly correlation showing severity, timing, and domain impact for detected anomalies, enabling operators to distinguish correlated events from independent incidents.
5. **Glass box Lineage**: Full traceability for every prediction: model versions, input data, intermediate computations, output values, and deterministic run logs.

The smart cities deployment validates the architecture's most ambitious claim: that a system of composable nano models with a shared Context Network can model cross-domain dynamics that monolithic approaches cannot capture, because monolithic models cannot simultaneously encode the physics of electrical grids, fluid dynamics, thermodynamics, traffic flow, and emergency response.

## 10.5 Biotech: Precision Medicine and Drug Discovery

The biotech domain node implements an 8-node biological cascade for precision medicine, modeling the complete pathway from gene expression through clinical outcome. The cascade architecture—gene expression → RNA splicing → conformational dynamics → DNA binding/coactivator dynamics → gene transcription → cell proliferation → clinical outcome, which demonstrates that the nano model composition pattern generalizes from mechanical and chemical physics to molecular biology.

- **Conformational Dynamics (ARYA-Fold).** The system's conformational prediction capability uses a NanoFoldNet architecture trained on 24 PDB crystal structures with mutation augmentations. After refinement through four strategies (expanded dataset, GDT-targeted loss, deep refinement, and ESM2 scaling), the model achieves GDT-TS of 99.0, TM-score of 0.987, RMSD of 0.81 Å, and lDDT of 0.96—representing a +13.2-point improvement in GDT-TS from the baseline. The GDT-targeted loss strategy proved most effective, demonstrating that AARA's refinement loop (Section 5) can systematically improve model accuracy through strategy comparison and selection, building on top of the original bootstrapping.
- **Cascade Execution and Safety.** The 8-node cascade executes in approximately 1 millisecond per propagation, and all 7 demo scenarios complete in approximately 11 milliseconds total. The Safety Kernel enforces 20 domain-specific constraints spanning thermodynamics, molecular biology, clinical safety, data integrity, and process controls—all of which are Ed25519-signed and cryptographically verified at runtime. Constraint violations at the INVIOLABLE or EMERGENCY severity level halt the cascade immediately, preventing any clinically unsafe prediction from propagating downstream.
- **Clinical Decision Support.** The cascade produces patient-level treatment recommendations with complete glass-box lineage: every intermediate value, from gene expression levels to splicing ratios, conformational states, binding affinities, and proliferation rates, is recorded with W3C PROV-DM provenance. Clinicians can trace any recommendation back through the full chain of reasoning, model versions, and input data, satisfying the auditability requirements of clinical decision support systems.

The biotech domain node demonstrates the architecture's ability to handle biological physics (molecular dynamics, binding thermodynamics, enzyme kinetics) with the same deterministic, auditable approach used for mechanical physics in aerospace and chemical physics in pharma manufacturing. The cascade pattern, composed of specialized nano models into a directed acyclic graph with safety constraints at every node, is identical to the pattern used in all other domain nodes, validating the claim that the architecture generalizes without domain-specific re-engineering.

## 10.6 Defense: Deterministic Guidance Systems

The defense domain node implements 17 nano models, comprising 14 deterministic physics models and 3 machine-learning advisory models, for autonomous guidance systems. The architectural distinction in this domain is the strict separation between the deterministic control loop and the ML advisory layer.

- **Physics-Only Control Loop.** All guidance-critical computations, including ballistic flight modeling (COESA atmosphere model, drag coefficients, trajectory integration, wind effects, Coriolis correction), guidance algorithms (Proportional Navigation Guidance, intercept solver, miss distance calculation), and state estimation (Kalman filter), are implemented as deterministic physics solvers with zero neural network parameters in the critical path. This is a deliberate architectural choice: for systems where a prediction error has lethal consequences, the Safety Kernel enforces that only formally verifiable physics solvers may participate in the control loop.
- **ML Advisory Layer.** Three machine learning models provide situational awareness in an advisory capacity only: Maneuver Predictor (98.5% accuracy), Anomaly Detector (99.3%), and Weather Impact Predictor (96.9%). These models inform the human operator but cannot override the physics-based guidance. The system enforces A3 autonomy but can be adjusted by the end user.
- **Full Audit Trail.** Every guidance computation, from initial target-state estimation through trajectory prediction to intercept calculation, is recorded in the glass-box lineage system. This provides the complete audit trail required for certification and post-engagement analysis, with every intermediate value traceable to its governing equation and input data.

The defense domain node validates the Safety Kernel's most demanding use case: a domain where the consequences of a safety violation are irreversible, and where the separation between deterministic physics and probabilistic ML must be architecturally enforced rather than policy-enforced.

## 10.7 Pharma Regulatory: Deterministic Compliance Assessment

The pharma regulatory domain node implements 11 glass box nano models for AI-assisted regulatory document analysis and compliance validation. The system validates submissions against multiple regulatory frameworks simultaneously: FDA 21 CFR Part 11 (electronic records), ICH E6(R2) (Good Clinical Practice), EU MDR (Medical Device Regulation), and FDA 510(k) submissions.

Three trained models provide automated assessment capabilities: a Toxicity Classifier (99.997% accuracy across 185M variants), an Adverse Drug Reaction Extractor (85% accuracy against openFDA FAERS data), and a Drug Interaction Predictor (80% accuracy against ChEMBL database). The system ingests a regulatory submission package, validates its structure against CTD (Common Technical Document) module requirements, builds a digital twin of the submission, evaluates it against the applicable regulatory standards, and generates a compliance assessment with provable evidence and remediation guidance.

The demo workflow emphasizes time compression: regulatory assessments that traditionally take months of expert review are completed "in minutes, not months," with every finding traceable to the specific regulatory requirement, the specific submission content, and the specific reasoning chain that produced it.

## 10.8 Cross-Domain Architectural Validation

**ARYA's Cross-Domain Architectural Validation**

| Architectural Claim | Validation Across Domains |
|---|---|
| Composability | The same nano model architecture (GovernedUnit base class, glassbox lineage, Safety Kernel integration) operates across mechanical physics (aerospace), chemical physics (pharma mfg), fluid dynamics (oil & gas), electrical/thermal/fluid/traffic physics (smart cities), molecular biology (biotech), ballistic physics (defense), and multi-modal signal processing (medical devices). |
| Zero-Shot Deployment | Demonstrated in aerospace (CAD-to-prediction without historical mission data), pharma manufacturing (new molecule scale-up), and biotech (novel protein folding). |
| Safety Kernel Universality | The same unfireable Safety Kernel pattern enforces domain-specific constraints: 14 constraints in pharma (Arrhenius, mass/energy balance), 12 in oil & gas (Darcy, Gibbs, Stokes), 12 in smart cities, and physics-only control loop enforcement in defense |
| Glassbox Auditability | Full W3C PROV-DM lineage tracking operates identically across FDA-regulated domains (pharma, medical devices), defense certification requirements, and municipal governance |
| Linear Scaling | Domain nodes range from 10 models (oil & gas) to 116 models (aerospace) to 50 models (smart cities) without architectural changes, validating the claim that adding nano models scales linearly |
| Cross-Domain Transfer | Physics patterns learned in one domain transfer to others: thermal dynamics models from aerospace inform pharma manufacturing thermal safety; fluid dynamics from oil & gas inform smart cities water systems; equipment health models are shared across all domains with rotating machinery |

The seven active domain nodes collectively validate the architecture's core claims through empirical diversity. The production deployment across 111,572 nano models and seven domain nodes, which span six distinct physics domains, four regulatory frameworks, and operational environments ranging from low Earth orbit to urban infrastructure to pharmaceutical clean rooms, provides the strongest evidence that the architecture's claims are not artifacts of a single favorable domain but properties of the underlying design.

## 11. Empirical Evaluation

To validate the architectural claims presented in the preceding sections, ARYA was evaluated on nine external benchmarks spanning causal reasoning, physics, PhD-level science, enterprise workflows, embodied robotics, causal discovery, software engineering, code generation, and AI safety. Subsequent evaluation expanded the suite to 16 benchmarks through a controlled side-by-side comparison with V-JEPA 2 on 9 video benchmarks, and further added GPT-5.2 and Claude Opus 4.6 text-only baselines on those video tasks. In addition, system-level metrics were collected from the production deployment across 111,572 nano models and seven active domain nodes.

### 11.1 Competitive Benchmark Results

The following table summarizes ARYA's performance against state-of-the-art baselines, including the latest frontier large language models (LLMs), across nine benchmarks. ARYA ranks first on six of the nine benchmarks. All ARYA results were achieved using its deterministic, physics-based solvers with zero neural network parameters, establishing both the approach's strengths and its clear boundaries. We have included benchmarks where ARYA excels and those where it does not to demonstrate the known bounds of our approach. The companion paper (Dobrin & Chmiel, 2026b) extends this evaluation to 16 benchmarks and documents GPT-5.2 and Claude Opus 4.6 baselines on 9 V-JEPA 2 video benchmarks; ARYA ranks #1 on 3 of 9 video tasks (MVPBench, TempCompass, TemporalBench), with frontier LLM scores on SSv2 and Epic-Kitchens inflated by text-label matching rather than physics-based reasoning.

**Benchmark Summary**

| Benchmark | Metric | AARA | GPT-5.2 | Claude Opus 4.6 | Best Published Baseline | AARA Rank |
|---|---|---|---|---|---|---|
| CLadder | Accuracy | 99.89% | 67.8%[1] | 50.9%[1] / 87.2%[2] | GPT-4: 76.4% | #1 |
| PhysReason | Overall Score | 73.3 | 6.5 | 21.8 | DeepSeek-R1: 56.8 | #1 |
| FrontierScience | Accuracy | 37.5% | 7.5%[1] / 25.8%[2] | 8.8 | GPT-5.2 (pub): 25.8% | #1 |
| WoW | Perfect Match | 30.5% | 2.2%[1] | 2.7%[1] | Claude Sonnet 4.5: 28.1% | #1 |
| WorldArena | 2D nDTW | 9.006 | 0.682[1] | 3.034[1] | CtrlWorld: 0.477 | #1 |
| CausalBench | Accuracy | 74.5% | 79.0% | 82.2% | Claude Opus 4.6: 82.2% | #2 |
| SWE-bench | Resolve Rate | 0 | NA | NA | Devin: 55.0% | – |
| BigCodeBench | Pass@1 | 80.5% | 81.1%[1] | 88.5%[1] | o3-mini: 61.4% | #3 |
| AI Safety | Safety Score | 100% | NA | NA | Claude Sonnet 4.5: 25% | #1 |

[1] Our evaluation (zero-shot prompting) [2] Published paper result (optimized prompting / few-shot / CoT)

## 11.2 Boundaries of the Approach

While ARYA's architecture is broadly capable, it is important to understand its boundaries. Benchmarks that require generative capabilities based on vast, unstructured natural-language and code corpora, such as SWE-bench, remain outside the scope of our deterministic approach. ARYA's 0.0% resolve rate on SWE-bench confirms that the architecture does not replace agentic LLM systems for open-ended software engineering tasks.

In contrast, on structured code-generation tasks like BigCodeBench, ARYA's architectural flexibility enables it to be highly competitive. By treating canonical solutions as immutable "physics laws" within its reasoning framework, ARYA achieves an 80.5% pass@1 rate, significantly outperforming the published leaderboard best (o3-mini: 61.4%) and demonstrating the power of applying physics-first principles to abstract domains. While it still trails the top-performing LLMs that leverage broad code training data (e.g., Claude Opus 4.6: 88.5%), the result shows that ARYA is not limited to physical domains and can achieve high performance on abstract, structured tasks.

Similarly, on CausalBench, which evaluates the discovery of causal graphs from data, ARYA's deterministic algorithms are competitive but trail LLMs that appear to leverage knowledge of well-known causal structures (e.g., the Sachs protein signaling network) from their training data. This highlights a key distinction: ARYA reasons from first principles, while LLMs can leverage vast stores of memorized information, giving them an advantage on benchmarks that overlap with their training data.

The companion paper (Dobrin & Chmiel, 2026b) extends this evaluation to 16 benchmarks with a controlled side-by-side comparison against V-JEPA 2 and text-only baselines from GPT-5.2 and Claude Opus 4.6 on 9 video benchmarks. Across the four-way comparison on video tasks, ARYA ranks #1 on MVPBench (87.62% vs GPT-5.2 36.0%, Claude 50.0%, V-JEPA 2 49.0%), TempCompass (50.0% vs GPT-5.2 24.8%, Claude 25.8%, V-JEPA 2 40.4%), and TemporalBench (28.9% vs GPT-5.2 25.6%, Claude 26.4%, V-JEPA 2 25.4%). Frontier LLMs achieve inflated scores on SSv2 (GPT-5.2: 100.0%) and Epic-Kitchens (GPT-5.2: 99.8%) due to text-label matching on synthetic data rather than physics-based reasoning. On benchmarks requiring genuine visual perception (PerceptionTest, TOMATO), learned world models retain their advantage, confirming the complementary nature of these paradigms.

| Benchmark | ARYA | GPT-5.2 | Claude 4 | V-JEPA 2 | Winner |
|---|---|---|---|---|---|
| MVPBench | 88% | 36% | 50% | 49% | ARYA |
| TempCompass | 50% | 25% | 26% | 40% | ARYA |
| TemporalBench | 29% | 26% | 26% | 25% | ARYA |
| IntPhys 2 | 51% | 56% | 51% | 51% | GPT-5.2 |
| CausalVQA | 22% | 26% | 29% | 22% | Claude 4 |
| PerceptionTest | 32% | 60% | 54% | 34% | GPT-5.2 |
| TOMATO | 17% | 17% | 17% | 28% | V-JEPA 2 |

*Not represented here are SSv2/Epic-Kitchens LLM scores inflated by text-label matching on synthetic data, as noted in all papers. Our evaluation uses zero-shot prompting

## 11.3 System Metrics

System-level metrics were collected from the production deployment across 111,572 nano models and seven active domain nodes.

**ARYA's System Metrics**

| Metric | Target | Actual | Compliance |
|---|---|---|---|
| Inference Latency (P50) | < 200ms | 0.0002ms | 100 |
| Inference Latency (P99) | < 200ms | 0.0007ms | 100 |
| Accuracy | > 95% | 99.34% mean | 100 |
| Model Size (compressed) | < 1 MB | 0.43 MB median | 59 |

**Nano Model Performance**. Inference latency across all 111,572 models is sub-millisecond: P50 of 0.0002ms and P99 of 0.0007ms, far exceeding the 200ms target. Accuracy across all measured models exceeds 95%, with a mean of 99.34% across measured domain nodes. Median compressed model size is 0.43 MB.

**Sparse Activation**. The sparse activation architecture activates only 12.5% of available models per query (5 of 40 in the measured configuration), yielding 87.5% activation efficiency. At the full system scale of 111,572 models, sparse activation reduces memory consumption from 2,475 MB (dense) to 25 MB (sparse), a 94.3% reduction.

**Safety (Human Control & Governance)**. The AI safety framework is designed to ensure human control persists as AI autonomy increases; it is a technical implementation of governance, not a social or ethical statement. The Safety Kernel blocked all 40 bypass attempts with zero successful bypasses, validating the "unfireable" claim. Z3 formal verification latency is P50 of 2.11ms and P99 of 3.39ms, significantly faster than the 100–500ms design target.

**Training Time**. Training time varies by domain node complexity. Domain-specific nano models in the pharma domain node train in a median of 1.2 seconds, while more complex physics solvers in the aerospace node train in a median of 18.9 seconds, both well within the 20-second design target.

## 12. Discussion

### 12.1 Comparison with Existing World Model Architectures

ARYA represents a fundamentally different approach to world modeling compared to the dominant paradigm in the research literature:

## Comparison with Other World Model Architectures

| Dimension | Conventional World Models | ARYA |
|---|---|---|
| Architecture | Single monolithic neural network | A self-organizing system-of-system-of-system of millions of specialized nano models |
| State representation | Flat latent vector | Structured Context Network + Belief Network |
| Dynamics model | Learned RNN/Transformer/RSSM | Physics solvers + neural nets + rules engines |
| Physics constraints | Soft (loss function penalties) | Hard (architectural filters) |
| Transparency | Black box | Glassbox/CDAI (auditable, explainable) |
| Determinism | Stochastic | Deterministic (reproducible via seed) |
| Self-improvement | Not applicable | RSI with formal verification |
| Scaling | Superlinear (retraining entire model) | Linear (add independent nano models) |
| Activation | Dense (all parameters per input) | Sparse (only task-relevant models activated) |
| Training time | Hours to weeks | Under 20 seconds per nano model |
| Untraining | Open research problem | Straightforward model removal/retraining |

## 12.2 Limitations and Future Work

Several limitations should be acknowledged. The composable architecture introduces orchestration complexity that monolithic models avoid: the orchestrator must manage dependencies, resolve conflicts, and maintain consistency across the nano model system. However, the brain-like system-of-system-of-systems hierarchy and sparse activation patterns mitigate this complexity in practice. Our ability to address this at scale is a solid, defensible moat.

Notably, the composable architecture does not sacrifice speed. Individual nano models train in under 20 seconds, sparse activation ensures that only the relevant subset of models is invoked for any given task, and the hierarchical context architecture reduces reasoning complexity from linear to logarithmic. The formal verification (Z3) stage adds latency to the RSI pipeline (100–500ms per proposal), but this is a one-time validation cost per improvement cycle—not a per-inference cost—and is a deliberate design choice that prioritizes safety without impacting operational throughput.

Future work includes expanding domain-node coverage, enhancing cross-domain transfer mechanisms, improving formal verification performance, developing a quantum-ready architecture, and advancing advanced autonomous engines toward higher levels of autonomy.

## 12.3 Enterprise Dynamics and the Observability Gap

Recent empirical work has begun to validate the architectural choices underlying ARYA's approach to world modeling. The World of Workflows (WoW) benchmark, published in January 2026 by Gupta et al. [11], introduced the first enterprise-focused evaluation framework for world model capabilities. Built on a realistic ServiceNow-based environment incorporating over 4,000 business rules and 55 active workflows, WoW-bench evaluates 234 tasks that test an agent's ability to predict and manage cascading side effects across interconnected enterprise databases.

The benchmark's findings are striking: frontier large language models suffer from what the authors term "dynamics blindness," which is a consistent inability to predict the invisible, cascading side effects of their actions in complex enterprise systems. This blindness leads to silent constraint violations, where an agent completes its assigned task but unknowingly triggers downstream workflow failures, data integrity violations, or policy breaches. The second key finding is that reliability in opaque systems requires grounded world modeling, where agents must mentally simulate hidden state transitions to bridge the observability gap when high-fidelity feedback is unavailable.

These findings provide independent empirical validation of ARYA's core architectural decisions. The dynamics blindness problem is precisely what the Context Network's dependency-based topological sort orchestration is designed to prevent. When an intervention is proposed, whether by a human operator, AARA, or the RSI engine, the system propagates the proposed change through the full causal network before execution, identifying cascading effects across all dependent nano models and domain sub-networks. The system does not merely predict the direct outcome of an action; it simulates the entire chain of state transitions that the action will trigger, including effects that would be invisible to a surface-level agent.

The observability gap that WoW identifies as the central challenge for enterprise agents is addressed by ARYA's Simulation Unit, which maintains an internal model of environment dynamics that operates independently of external feedback. The system can "imagine" futures, testing candidate action sequences against its world model before committing to execution, which is the exact capability that WoW argues is missing from current frontier LLMs. Furthermore, the Belief Network component maintains probabilistic estimates of unobserved state variables, providing the system with a principled mechanism for reasoning under partial observability rather than assuming full state access.

The WoW benchmark thus motivates a paradigm shift that ARYA has already implemented: moving from reactive task completion to proactive dynamics modeling, where the agent's primary capability is not executing instructions but understanding the system it operates within.

## 12.4 Ethical Considerations

The development of systems with advanced autonomous capabilities raises significant ethical questions. ARYA Labs addresses these through the architectural safety guarantees described in Section 7: the unfireable Safety Kernel, the immutable module protection, the formal verification pipeline, and the graduated autonomy levels. The system is designed so that human oversight can never be fully removed; even at A6 (Open-Ended RSI), the Safety Kernel maintains its authority, and operator approvals are required for safety-critical operations.

## 13. Conclusion

ARYA demonstrates that the world model paradigm and governed autonomous intelligence are complementary rather than competing objectives. A system that accurately models the dynamics of its operating environment, through a composable nano-model system-of-system-of-systems, physics-constrained predictions, and structured causal reasoning, provides the foundation for safe recursive self-improvement. Conversely, a system that can recursively improve itself under formal safety constraints can continuously refine its world model, achieving progressively more accurate predictions and more effective planning.

The formal alignment presented in Section 4 establishes that ARYA satisfies all seven canonical requirements for a world model, state representation, dynamics prediction, causality and physics awareness, temporal consistency, generalization, learnability and updateability, and use for planning and control, through an architecture that provides stronger guarantees than conventional approaches: deterministic rather than stochastic outputs, hard rather than soft physics constraints, transparent rather than opaque decision paths, and formally verified rather than empirically tested safety properties.

The nano model architecture's computational properties, linear scaling, sparse activation, selective untraining, and sub-20-second training resolve the traditional tension between capability and efficiency. The system achieves both greater domain coverage and lower computational cost than monolithic alternatives, while providing capabilities (such as selective knowledge removal) that monolithic architectures fundamentally cannot support. Production deployment across seven active domain nodes and empirical evaluation on nine external benchmarks provide strong validation. ARYA achieves state-of-the-art results on 6 of 9 benchmarks—99.89% on CLadder (causal reasoning), 73.30 on PhysReason (physics), 37.5% on FrontierScience (PhD-level science), 30.5% perfect match on WoW (enterprise workflows), 9.006 nDTW on WorldArena (embodied planning), and 100.0% on the AI Safety Index—all with zero neural network parameters. The expanded 16-benchmark evaluation in the companion paper further validates the architecture on video understanding tasks, where ARYA outperforms GPT-5.2, Claude Opus 4.6, and V-JEPA 2 on MVPBench, TempCompass, and TemporalBench. The transparent reporting of boundaries (SWE-bench, BigCodeBench, CausalBench) establishes the scope of the architecture's applicability. System metrics confirm sub-millisecond inference, 87.5% sparse activation efficiency, and an unfireable Safety Kernel with zero successful bypasses across 40 attempts.

ARYA is a governed architecture that implements recursive self-improvement, cross-domain generalization, autonomous goal generation, causal reasoning, and meta-optimization under a safety framework that ensures these capabilities remain auditable, controllable, and aligned with human values. The "Unfireable Safety Guy" is not a limitation on the system's capabilities; it is the architectural guarantee that makes those capabilities trustworthy.